\documentclass[review]{elsarticle}
\usepackage{amsmath,amssymb,amsfonts}
\usepackage{graphicx}
\usepackage{textcomp}
\usepackage{algorithm}
\usepackage{algorithmic}
\usepackage{multirow}
\usepackage{url}

\usepackage{hyperref}
\usepackage{geometry}
\journal{Journal of \LaTeX\ Templates}

\begin{document}
	
	\begin{frontmatter}
		
		\title{Evolving Large-Scale Data Stream Analytics based on Scalable
			PANFIS}

		\author{Mahardhika Pratama\fnref{myfootnote}}
		\author{Choiru Za'in\fnref{myfootnot}}
		\author{Eric Pardede\fnref{myfootnot}}
		\address{}
		\address{}
		\fntext[myfootnote]{Nanyang Technological University, Singapore}
		\fntext[myfootnot]{La Trobe University, Australia}
		
		
		\address[mysecondaryaddress]{Nanyang Technological University
			50 Nanyang Avenue, Singapore 639798}
		\address[mymainaddress]{Plenty Rd \& Kingsbury Dr, Bundoora VIC 3086, Australia}

		\begin{abstract}
			The main challenge in large-scale data stream analytics lies in the ability of machine learning to generate large-scale data knowledge in reasonable timeframe without suffering from a loss of accuracy. Many distributed machine learning frameworks have recently been built
			to speed up the large-scale data learning process. However, most distributed machine learning used in these frameworks still uses an offline algorithm model which cannot cope with the data stream problems. In fact, large-scale data
			are mostly generated by the non-stationary data stream where its pattern
			evolves over time. To address this problem, we propose a novel Evolving
			Large-scale Data Stream Analytics framework based on a Scalable Parsimonious Network based on Fuzzy Inference System (Scalable PANFIS), where the PANFIS evolving algorithm is distributed over the worker nodes in the cloud to learn large-scale data stream. Scalable PANFIS framework incorporates the active learning (AL) strategy and two model fusion methods. The AL accelerates the distributed learning process to generate an initial evolving large-scale data stream model (initial model), whereas the two model fusion methods aggregate an initial model to generate the final model. The final model represents the update of current large-scale data knowledge which can be used to infer future data. Extensive experiments on this framework are validated by measuring the accuracy and running time of four combinations of Scalable PANFIS and other Spark-based built in algorithms. The results indicate that Scalable PANFIS with AL improves the training time to be almost two times faster than Scalable PANFIS without AL. The results also show both rule merging and the voting mechanisms yield similar accuracy in general among Scalable PANFIS algorithms and they are generally better than Spark-based algorithms. In terms of running time, the Scalable PANFIS training time outperforms all Spark-based algorithms when classifying numerous benchmark datasets.		
		\end{abstract}
		
		
	\end{frontmatter}

	\section{Introduction}\label{sec:introduction}
	
	Large-scale data stream analytics has become one of the emerging area in data science \cite{chen2012business,fernandez2014big,chen2014data}.
	A large volume of data in many forms (e.g. text, picture, sound, video,
	signals etc.) can be generated from numerous sources (e.g. IoT, Web
	2.0, and social networks) in the Internet era. These information are essential for many companies/corporations to support their urgent decision-making to ensure their competitive advantage.
	
	Extracting valuable knowledge from large-scale data stream is challenging due to its 4V characteristics: volume, velocity, variety and veracity. Large-scale data stream is mostly  generated  by real-world applications in which data arrive continuously in non-stationary environments. With its characteristic, the velocity, 	
	it is important to obtain knowledge from large-scale data efficiently (reasonable timeframe without a reduction in the algorithm’s accuracy) \cite{ramirez2018big}. 
	
	Large-scale data stream analytics problem can be solved by two ways: 1) distributed computing ; and 2) streaming algorithm \cite{bifet2014big}. Distributed computing focuses on how to distribute/parallelize data processing from a single-node CPU-based processing into multi-node cluster-based processing framework \cite{hu2014toward}, thus accelerate the learning time. Streaming algorithm also known as evolving algorithm processes/learns data at high speed, single pass, and online manner. Its structure is evolving following an update of the current datum. It does not require historical data as the current information/pattern/model is
	discarded after the last datum has been learned. This feature can help to reduce the storage requirement because the historical data do not need to be retained.
	
	Recent work on large-scale data analytics was reported in \cite{del2015mapreduce},  utilizing the MapReduce \cite{dean2008mapreduce} method. 
	In this work, the distributed algorithm used to learn the data partition was still based on the offline algorithm, namely Fuzzy Rule Based classifiers (FRBCSs) \cite{ishibuchi2004classification} to model complex problems. However, the offline learning is not efficient especially in handling rapid varying and a vast amount of large-scale data stream. 
	On the other hand, processing large-scale data using a single-node evolving algorithm is limited by the memory and bandwidth of a single machine. This issue remains the main challenge for further developments in large-scale data analytics. Taking both benefits of distributed processing and online data processing, the large scale data stream analytics framework should accommodate between scalability of distributed learning and the efficiency of evolving algorithm. 
	
	In this work, we propose an Evolving large-scale data stream analytics framework based on Scalable PANFIS, where PANFIS \cite{pratama2014panfis} is a seminal evolving algorithm based on a hybrid neuro-fuzzy system which has the capability to learn the data stream in the single pass mode to cope with the high speed and dynamically changing data stream. The three methods are involved in the Scalable PANFIS framework: 1) active learning (AL); 2) the rule merging; and 3) the majority voting.
	
	The training phase of this framework is conducted by distributing the PANFIS algorithm (with or without AL) across the worker nodes. AL is the method to accelerate the learning process by selecting the important instance of training data. Fig. \ref{Figure1} in the blue box part illustrates that PANFIS (with or without AL) learns data stream partition in the worker node. 
	Furthermore, the rule merging is designed to aggregate several models from different data stream partitions to yield single model for inference task. The majority voting method is applied to acquire the output from the majority decisions conducted by multiple classifiers in the system.
	
	To summarize, the main contributions of this work are as follows:
	\begin{itemize}
		\item 
		We present Scalable PANFIS framework, an Evolving large-scale data stream analytics framework, a distributed streaming algorithm, which can deal with large-scale data stream prediction problems. This framework is scalable/distributable as well as it can cope with the dynamic and evolving data stream.
		\item Our evolving large-scale data stream analytics framework is developed under four structures using combinations of active learning, rule merging and voting mechanisms. The four structures are capable of processing large-scale data stream in the one-pass fashion.	
		\item 
		We present the robust rule merging method to solve the aggregation problem in large-scale distributed data stream training. Because different data partitions are fed to the worker nodes and possibly distracts the original data distribution, the rules generated from the training data partitions (initial model) could not be applied directly to merge different model of data partitions. This issue is mainly due to poor rules generated using different tendency of original data distribution. Our rule merging method can solve the aggregation problem and yields the stable performance (accuracy) for all dataset.

	\end{itemize}
	The rest of the paper is organized as follows. Section 2 discusses
	the related research: PANFIS' algorithm and learning
	policy. Section 3 explains the architecture of the large-scale data analytics framework.
	Section 4 describes our proposed approach which specifically
	explains the active learning strategy and two aggregating methods.
	Section 5 discusses the numerical study: experimental
	setup, algorithms comparisons, and the numerical results
	of large-scale data analytics. Section 5 presents the discussion
	and section 6 concludes the paper.
	
	\section{PANFIS}	
	
	PANFIS is an evolving algorithm which is built on evolving neuro-fuzzy systems (ENFS), an extension of the well-known classic neuro-fuzzy systems(NFSs) \cite{lin1996neural}. NFSs combine fuzzy systems which follow the principle of human reasoning
	and neural networks which have a learning ability, parallelism, and
	robustness characteristics. Basically, ENFSs are the evolving version
	of NFSs and have the capability to evolve their structure (rules) so that
	they can adapt to the changing environment, which is essential to cope with the non-stationary environment.
	
	The PANFIS evolving algorithm can learn the data without an initial structure.
	During the learning, its structure (fuzzy rules and its parameters) is evolving, so that the 
	new rule can be generated, updated, and pruned. In addition, the the merging process is carried out
	by identifying identical (or similar) fuzzy rule sets to simplify the rules complexity.
	
	The main feature of PANFIS is the construction of ellipsoids in arbitrary position to support multidimensional membership function in the feature space. Although the inference process of PANFIS is carried out in the high-dimensional space, this ellipsoids in arbitrary position can be projected to fuzzy sets to form the antecedent parts of fuzzy sets which is interpretable for the user using two fuzzy set extraction methods.
	
	The evolution of rules is controlled by the datum significance (DS)
	criterion which represents potential of current datum being learned
	in the system. DS was initially proposed by \cite{huang2005generalized} and \cite{rong2006sequential} to identify
	the significance of a datum which can be measured by its statistical contribution to PANFIS' output. Once, its value is high, this datum is considered to have high descriptive power and is thus a good candidate as a new rule.
	
	The rule adaptation policy is executed when the arrival datum falls in the current clusters. In this case, the winning rule parameters are adjusted to determine the new coverage/span of the winning rule. Rule adaption in the original PANFIS utilize ESOM. However, this method has the drawback of instability which requires re-inversion once the inverse covariance matrix is ill-conditioned (e.g., due to redundant input features). As a result, the adaptation formula of GENEFIS \cite{pratama2014genefis}, pClass \cite{pratama2015pclass}, and GEN-SMART-EFS \cite{lughofer2015generalized} is adopted instead.
	
	The three properties of the winning rule are updated as follows:
	\begin{equation}
	C_{win}^{update}=\frac{N_{win}^{prev}}{N_{win}^{prev}+1}+\frac{X-C_{win}^{prev}}{N_{win}^{prev}+1},
	\end{equation}
	\begin{flalign}
	\nonumber
	\textstyle \sum_{win}(update)^{-1} &= \frac{\sum_{win}(prev)^{-1}}{1-\alpha}+\frac{\alpha}{1-\alpha},&&
	\end{flalign}
	\begin{flalign}
	\textstyle
	\frac{\sum_{win}(prev)^{-1}(X-C_{win}^{update}))    (\sum_{win}(prev)^{-1}(X-C_{win}^{update}))^T}
	{1+\alpha(X-C_{win}^{prev})\sum_{win}(prev)^{-1}(X-C_{win}^{prev})^T},&&
	\end{flalign}
	\begin{equation}
	N_{win}^{update}=N{win}^{prev}+1,
	\end{equation}
	where $	C_{win}^{update}$, $\sum_{win}(update)^{-1}$, and $N_{win}^{update}$ denote the updated focal point, dispersion matrix, and the population of the winning rule.
	
	Rule pruning of PANFIS is driven by an extended rule significance
	(ERS) concept which represents the contribution of every rule to
	the system output. ERS is inspired by the concept in the SAFIS method
	\cite{rong2006sequential} but is extended by integrating hyperplane consequents
	and generalizing to ellipsoids in an arbitrary position which enable
	rules to be pruned in the high-dimensional learning space.
	
	The fuzzy sets merging in PANFIS is carried out when some fuzzy sets are overlapped, which means
	they have similar membership function. This is done to reduce fuzzy rule redundancies, thus forming
	interpretable rule base. The similarity calculation between two fuzzy sets can be found in \cite{lughofer2011line}. The two fuzzy sets can be merged if the similarity $ Sker\geq0.8$. Note that this mechanism is carried out only if one wish to produce classical interpretable rule with t-norm operators as AND part of the rules. This step is required because the projection of high-dimensional rule to one dimensional space often results in the overlapping issue. 
	
	The fuzzy consequences adjustment of PANFIS is driven by the enhanced recursive least squares (ERLS), which is
	inspired by conventional least squares (RLS) \cite{wittenmark1995adaptive}. The main function of ERLS is designed to support the convergence of the system error, which is used for weight vector updates.

	\section{Four Structures of Scalable PANFIS for Evolving Large-Scale Data Stream Analytics}
	Scalable PANFIS framework covers three major phases: training phase, aggregating phase, and testing phase. In this section, we divide the discussion into two parts. The first part (subsection \ref{subsec:ScalablePanfisArch}) discusses the Scalable PANFIS training mechanism which specify on how PANFIS (with or without AL) is parallelized in the Apache Spark (Spark) platform \cite{zaharia2010spark}. At the end of training process, the driver node receives the results obtained by learning of PANFIS for each data stream partitions as an initial model. The second part (subsection \ref{subsec:StructureLearning}) discusses about how the initial model is processed further through aggregation mechanism to deliver the final knowledge base or model.

	\subsection{Scalable PANFIS Framework}\label{subsec:ScalablePanfisArch}
	Apache Spark platform \cite{zaharia2010spark} is regarded as the latest platform for distributed-based data processing. In comparison with the older platform, such as Hadoop, Spark improves performance significantly in terms of speed in data processing because it supports an in-memory based instead of disk-based programming model.
	The	Spark ecosystem contains two parts: 1) spark-core ; and 2) programming interface core. Spark-core lies in the lower level library of the Spark ecosystem to serve the programming interface core. The programming interface core is integrated by Spark APIs which support many programming languages such as
	Scala, Java, Python, and R. Furthermore, Spark API also provides a machine learning library (Spark MLib), GraphX for analysis, a stream processing module of Spark Streaming, and SQL for structured data processing. For large-scale data analytics, these Spark ecosystem components enable framework to conduct parallel data processing and support the real-time insight/knowledge generation of large-scale data stream.
	
\begin{figure}[htbp]
		\begin{centering}
			\includegraphics[width=9cm]{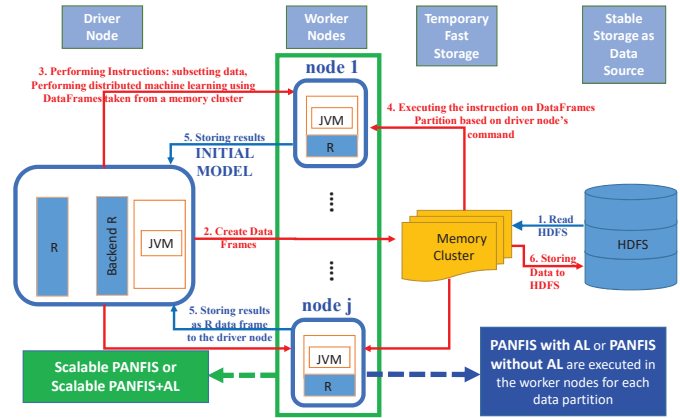}
			\par\end{centering}
		\caption{The Data Flow Architecture Scalable PANFIS Framework during the data stream training phase in Spark Platform\label{Figure1}}
	\end{figure}

	The R language is chosen as the main programming language in Scalable PANFIS framework as it is a well known programming language for data analysis. In order to bridge the operation between R and Spark, SparkR library (as a backend R and Java Virtual Machine (JVM)) is utilized to manipulate and process large-scale data in a parallel/distributed manner.
	The type of data used in processing large-scale data in parallel mode is Spark DataFrames (DataFrames), a unique Spark
	data abstraction which is stored in memory cluster computing. The diagram in Fig. \ref{Figure1} shows the data flow architecture of Scalable PANFIS framework in the training phase utilizing Spark platform.

	The Spark platform consists of a driver node and some worker nodes along with the data sources (e.g, Hadoop Distributed File System (HDFS)). The training process of Scalable PANFIS framework operation is initialized by acquiring the dataset from the data source which is shown in step 1 Fig. \ref{Figure1}. The second step creates the DataFrames in the memory cluster, instructed by the driver node, as a result of a data loading process from the HDFS data source to be used in many future Spark operations such as querying, subsetting, feeding the data into machine learning, etc. In our Scalable PANFIS framework, after the DataFrames are created in the memory cluster, they can be manipulated repetitively. For example, DataFrames can be partitioned and distributed to be processed in the worker nodes as shown in step 3. Then, the partitioned/chunked data are processed in the worker nodes by using either the Spark machine learning library or the user-defined function (e.g. PANFIS with AL) as shown in step 4 in Fig. \ref{Figure1} (see the blue solid box with the dashed arrow). The results/models are then sent back to the driver node as an initial model in step 5.
	Step 6, is an optional step depending on whether the models should be saved into the stable storage for back up purposes or directly used for the next process (rule merging and majority voting). Please note that step 5 is the end of the Scalable PANFIS framework's training process (initial model is generated in the driver node). The initial model is then further aggregated into the final model, whose process is depicted using either the rule merging and majority voting methods.
	
	\subsection{Structure of Scalable PANFIS Framework model}\label{subsec:StructureLearning}
	
	This subsection details the four structures of Scalable PANFIS framework. The first and second structure utilize Scalable PANFIS (without AL) in generating initial model. While the first structure uses the rule merging, the second structure utilizes the majority voting in aggregating the initial model. The third and the fourth structure uses the first two structures added with AL. 
	
	\subsubsection{Scalable PANFIS Framework using the Rule Merging Method}\label{subsubsec:SCPANFISMerging}
	
\begin{figure}[htbp]
		\begin{centering}
			\includegraphics[width=9cm]{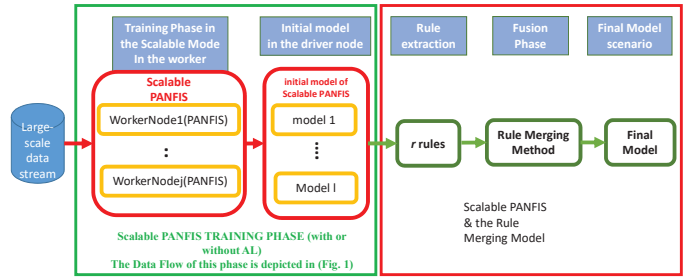}
			\par\end{centering}
		\caption{The Structure of Scalable PANFIS Framework using the Rule Merging Method\label{Structure13}}
	\end{figure}

	This structure of Scalable PANFIS framework uses distributed PANFIS to train large-scale data stream in generating the initial model. The initial model is then aggregated using the rule merging method to generate the final model. This structure is depicted in Fig. \ref{Structure13}.
	
	The initial model of Scalable PANFIS framework contains many classifiers with many models generated (initial model) in the training phase. The initial model contains $l$ models resulted from $l$ data stream partitions training. From $l$ models, $r$ rules are extracted prior to rules merging as depicted in Fig. \ref{Structure13}, assuming that each model contains one or more rules, so that $r\geq l$.

	Unlike the single classifier system, where rules can be used directly to test the testing dataset, using all $r$ rules (extracted from initial model) directly to infer the testing dataset is impractical because the classification system contains many rules which are overlapping to each other. The overcomplex rules may deteriorate the generalization power of the classifier because they are generated by different subsets of a big dataset where some of them may be poor due to low supports or crafted by a small picture of the true distribution. To overcome this drawback, The rule merging method is applied to simplify the classifier's complexity.
	
	Rule merging method has previously been applied in the big data processing with fuzzy system such as \cite{zabig} and \cite{za2018big}. However, applying the rules merging method directly results in performance degradation.
	We investigate that the decrease in accuracy is caused by the class overlapping problem. Furthermore, outliers may appear and may generate new rules having a very small number of populations. To validate our hypothesis, we conduct an experiment on HEPMASS dataset to show that the initial rules selection (selecting $k$ best rules) and rules removal (eliminating rules which have a small population) prior to rules merging influence the classifier performance. The results of this experiment is illustrated in the Table \ref{HepmassTesting}.	
	\begin{table}[htbp]%Average Accuracy
		\centering
		\caption{The accuracy of HEPMASS testing dataset for different $k$ best initial rules selection with and without rule removal prior to rules merging}\label{HepmassTesting}
		\scriptsize
		\begin{tabular}{|c|c|c|}
			\hline
			\textbf{k}&\textbf{Accuracy(rules removal)}&\textbf{Accuracy(without rules removal)}\\
			\hline
			1&83.87&83.57\\
			\hline
			3&83.47&83.68\\
			\hline
			5&83.15&83.37\\
			\hline
			45&83.63&83.46\\
			\hline
			50&83.52&82.82\\
			\hline
			55&77.28&62.16\\
			\hline
			60&71.11&61.64\\
			\hline
		\end{tabular}
	\end{table}
	
%	It can be seen from the empirical result in Table \ref{HepmassTesting} that not all rules in the classification system provide the same classification output. It is clear that at $k=55$, the performance of HEPMASS testing dataset decreases with the higher rate on without rules removal. For the case of rules removal, there are no outliers involved, the performance decrease purely due to the existence of rules which have a low generalization power. For the case of without rules removal, the number of rules which have a low generalization power is higher than the classifier with rules removal, thus performs worse result. This result indicates that some outliers are wrongly chosen because they have higher ranking than other rules. (OLD VERSION)
It can be seen from the empirical result in Table \ref{HepmassTesting} that outliers appear as indicated in the classification results and decrease the classification performance. The results in Table \ref{HepmassTesting} (Accuracy without rules removal) shows that whenever some rules from the outliers are mistakenly involved in the rule merging process. The accuracy decreases  from around 82 percent at $k=55$ to around 62 percent. On the other hand, even though some outliers have been removed, in Table \ref{HepmassTesting} (Accuracy with rules removal) the classification performance also decrease with a lower level than without rules removal. Therefore, we conclude that we should carefully design the rule merging method/mechanism by choosing the optimum number of rules (only the $k$ highest confidence rule) and by removing the outliers prior to rule merging process in order to yield the high classification result.

	 The rules selection is inspired by the work in \cite{del2015mapreduce}, where they select the rules with the highest weight which among the same antecedent in the same partition. The same procedure is also repeated, where the selected rules from every partition  are compared with other selected rules from other partitions. Rules which have a highest weight among the same antecedent will be selected. At the end of the process, only the clusters/rules which have the highest weight for all unique antecedents will exist and will be be used for the final large-scale data modeling.
	
	In this framework, the $k$ best models are selected by observing the highest classification accuracy among training data partitions. The training accuracy reflects the confidence level of the model to be recruited as the base model. Since each model is constructed by one or more fuzzy rules, the weight of rules are assigned by the weight of its corresponding model determined from the training accuracy.
	
	Suppose that the data stream is denoted as $ds=\{x_1,x_2,x_3,..,x_n,...\}$ in the feature space $R^2$, where $n$ shows the index of $n$-th datum (current datum). Thus, $n$ number of training data will be fed to one worker node (training set) denoted as $S$. Thus, for $l$ data partitions in the training set, the collection $l$ partitions of the training set and its corresponding models $M$ respectively can be constructed as:
	\begin{flalign}
	\textstyle S_{Train}=\{S_1,S_2,...,S_l\}\\
	\textstyle M_{Train}=\{M_1,M_2,...,M_l\}.
	\end{flalign}

	Prior to the merging process, several steps need to be carried out to process initial model: 1) extracting all rules from all models; 2) assigning the weight of $r$ rules, where rule weight can be obtained from the weight of the model corresponding to it; 3) eliminating the rules which have a very small population; 4) selecting the $k$ best rules which have the highest classification accuracy.
	
	\begin{figure}[htbp]
		\begin{centering}
			\includegraphics[width=16cm]{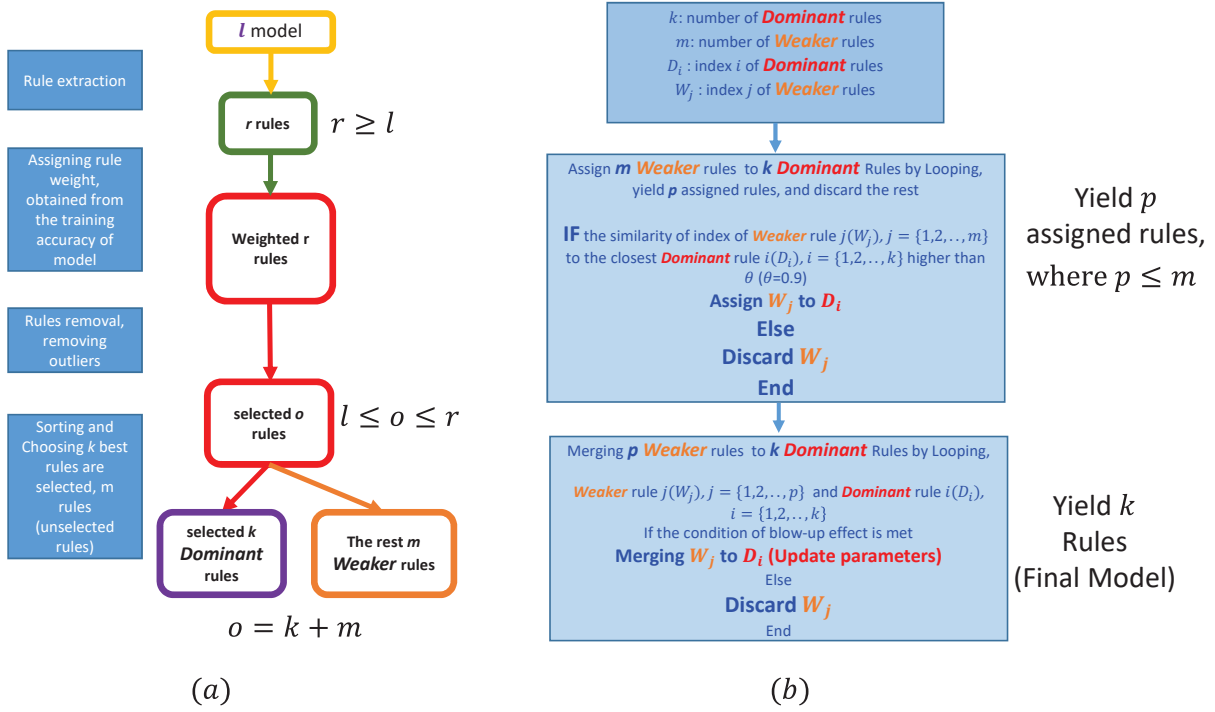}
			\par\end{centering}
		\caption{(a) The preliminary steps prior to merging in visual representation (b) The Merging Process in visual representation based on Algorithm\ref{merging} \label{baganMerging}}
	\end{figure}

	The preliminary steps of the merging process is carried out by simply concatenating all rules in the system, resulting in $r$ rules being extracted from $l$ models, $r \geq l$. The next step is to assign the rule weight which can be acquired from the training accuracy of the model. The rules removal in step three aims to remove the poor rules due to outliers. The effect of outliers can be observed if rules have low populations, thus they contribute less during their lifespan.
	We apply 5 percent threshold population of the rule in the total population of the cluster. If it does not meet the requirement of the threshold, the rule is removed from the system. From this step, $o$ number of rules are extracted, where $l \leq o \leq r$. Up to this step, we have $o$ rules candidate to be fed in the merging process.
	
	The merging process is initialized by choosing the most $k$ influential rules among other $o$ rule candidates. We call the $k$-most influential rules the \textbf{Dominant} rules, whereas the $m$ number of rules are called the \textbf{Weaker} rules thus, $o=m+k$. If the set of \textbf{Dominant} rules is denoted by $D_j=\{D_1,..,D_j,...,D_k\}$, and the  \textbf{Weaker} rules is denoted by $W_i=\{W_1,..,W_i,...,W_m\}$. The merging process between rules occurs by following the procedure illustrated in algorithm \ref{merging}. Both preliminary steps of merging and the merging process (Algorithm \ref{merging}) are visualized in Fig. \ref{baganMerging}.
	
	\begin{algorithm}[!t]
		\caption{Rule Merging Algorithm}
		\label{merging}
		\algsetup{linenosize=\tiny}
		\scriptsize
		\begin{algorithmic}
			\STATE \textbf{Input} : Set of Dominant and Weaker Rules (D and W)
			\STATE \textbf{Onput} : Set of Updated parameter of Dominant Rules
			\\  \textit{Initialization}:
			\\ $k$ : number of \textbf{Dominant} rules
			\\ $m$ : number of \textbf{Weaker} rules
			\\ $D_i$ : index $i$ of \textbf{Dominant} rules
			\\ $W_j$ : index $j$ of \textbf{Weaker} rules
			\\  \textit{Loop Process}:
			\\ -Assign Weaker Rules to the closest Dominant Rules
			\FOR {$j = 1$ to $m$}
			\FOR {$i = 1$ to $k$}
			\STATE Count \textbf{Similarity} between $W_j$ and $D_i$ ($Sim_{D_i,W_j}$) (\textbf{(Formula \ref{eq:FormulaCriterion2})})
			\ENDFOR
			\STATE Determine winner rule: calculating maximum similarity
			\STATE \textbf{$D_i(winner)$ }$=\arg\max_{j=1,..,k} (Sim_{D_i,W_j})$
			\IF{(\textbf{$Sim_{D_i,W_j}(winner)$} $\geq$ $\theta$ ) \textbf{(Formula \ref{eq:FormulaCriterion2})}}
			\STATE Rule $W_j$ is recruited in the rule $D_i$
			\ELSE
			\STATE Discard $W_j$ (\textit{$W_j$ is regarded has a low similarity over current \textbf{Dominant} rules})
			\ENDIF
			\ENDFOR\\
			Some rules may be eliminated, resulted in $p$ assigned rules to the next merging process.
			\\  \textit{Loop Process}:
			\\ Merging Of Assigned Weaker Rules to the closest Dominant Rules
			\FOR {$i = 1$ to $k$}
			\FOR {$j = 1$ to $p$}
			\STATE
			- Iteratively update Dominant rule parameters by merging it with the assigned list of Weaker rules\\
			\IF{(\textbf{$j$} not in the list \textbf{$D_i$})}
			\STATE skip for the next $D_i$
			\ELSE

			\IF{( the condition of blow-up effect is met- \textbf{Formula (\ref{eq:FormulaVmerged})} )}
			\STATE
			$D_i(Update)$=Merging of rule $D_i(current)$ and rule $W_j$
			\textbf{ (Formula \ref{newcenter} \ref{newspread} \ref{newpop} \ref{newweight})}
			\\
			$D_i(Current) = D_i(Update)$\\
			\ELSE
			\STATE
			Cancel merging, discard rule $W_j$
			\ENDIF
			\ENDIF
			\ENDFOR
			\ENDFOR
		\end{algorithmic}
	\end{algorithm}

	The selected $k$- \textbf{Dominant} rules become the reference of the other $m$ number of \textbf{Weaker} rules, assuming that \textbf{Dominant} rules have better results than the \textbf{Weaker} ones. From Algorithm \ref{merging}, the first loop aims to assign each \textbf{Weaker} rules to the closest \textbf{Dominant} rule as well as discard the rule if the maximum similarity obtained less or equal than $\theta$ , where $\theta$ is set to 0.9 in this experiment, assuming that the higher the level of threshold, the more similar rules will be merged to avoid the decrease of classification performance. The result is the
	$p$ number of rules are assigned, where $p \leq m$ assuming that some rules are discarded.
	
	The similarity calculation between two rules is adopted from the method proposed in \cite{lughofer2015generalized}.
	This is based on the degree of deviation in the hyper-planes's gradient information, where the deviation is calculated based on the dihedral angle of the two hyper-planes they span, which is calculated as follows:
	\begin{equation}
	\phi=arccos(\frac{a^{T}b}{|a||b|}),\label{eq:FormulaAcos}
	\end{equation}
	where $a=(D_{i;1}D_{i;2}D_{i;3}-1)^{T}$ and $b=(W_{j;1}W_{j;2}W_{j;3}+1)^{T}$
	the normal vectors of the two planes corresponding to rules $D_i$
	and $W_j$, showing into the opposite direction with respect to target
	$y$ (-1 and +1 in the last coordinate). Thus, the similarity of two hyper-planes
	is formulated as follows:
	\begin{equation}
	Sim_{(D_{i},W_{j})}=\frac{\phi}{\pi},\label{eq:FormulaCriterion2}
	\end{equation}
	Two hyper-planes are regarded as similar if the similarity degree $Sim_{(D_{i},W_{j})} \geq \theta$.
	
	The next loop in Algorithm \ref{merging} performs the merging of $p$ assigned rules to the associated \textbf{Dominant} rules. The parameter update of the dominant rules is carried out iteratively, where \textbf{Dominant} rules can only be updated with the list of the \textbf{Weaker} rules assigned to them. However, before performing the merging process where the value of both consequent and antecedent
	parameters between the \textbf{Dominant} rule and the \textbf{Weaker} rule are merged, another criterion also should be met to ensure that both merged rules should form a homogeneous shape and direction. It also to represents the accurate representation of the two rules/clusters. Thus, the blow-up effect is applied to trigger homogeneous joint regions. The blow up effect is formulated as:
	\begin{equation}
	V_{merged}\leqslant n(V_{Dominant_i}+V_{Weaker_j}),\label{eq:FormulaVmerged}
	\end{equation}
	where $V$ stands for volume and $n$ is the dimension of input attribute.
	After these two conditions are fulfilled (formula \ref{eq:FormulaCriterion2} and formula \ref{eq:FormulaVmerged}), the updating
	parameters of merged rule is referring to the weighted average principle as applied in \cite{pratama2014genefis} as follows:
	\begin{equation}
	\begin{array}{c}
	c_{Dom_i}^{update}=\frac{c_{Dom_i}^{cur}N_{Dom_i}^{cur}+c_{Weak_i}^{cur}N_{Weak_i}^{cur}}{N_{Dom_i}^{cur}+N_{Weak_i}^{cur}},
	\end{array}\label{newcenter}
	\end{equation}
	\begin{equation}
	\begin{array}{c}
	\sum_{Dom_i}^{-1}(update)=\frac{\sum_{Dom_i}^{-1}(cur)*N_{Dom_i}^{cur}+\sum_{Weak_i}^{-1}(cur)*N_{Weak_i}^{cur}}{N_{Dom_i}^{cur}+N_{Weak_i}^{cur}},
	\end{array}\label{newspread}
	\end{equation}
	\begin{equation}
	\begin{array}{c}
	N_{Dom_i}^{update}=N_{Dom_i}^{cur}+N_{Weak_i}^{cur}
	\end{array}\label{newpop},
	\end{equation}
	\begin{equation}
	\begin{array}{c}
	w_{Dom_i}^{update}=\frac{w_{Dom_i}^{cur}*N_{Dom_i}^{cur}+w_{Weak_i}^{cur}*N_{Weak_i}^{cur}}{N_{Dom_i}^{cur}+N_{Weak_i}^{cur}}
	\end{array}\label{newweight},
	\end{equation}
	where $c_{Dom_i}^{update}$ and $\sum_{Dom_i}^{-1}(update)$ are the updated antecedent parameters of the merged rule and $w_{Dom_i}^{update}$ is the updated consequent parameter of the merged rule.

	\subsubsection{Scalable PANFIS Framework using the Majority Voting Method}\label{subsubsec:SCPANFISVoting}
\begin{figure}[htbp]
		\begin{centering}
			\includegraphics[width=9cm]{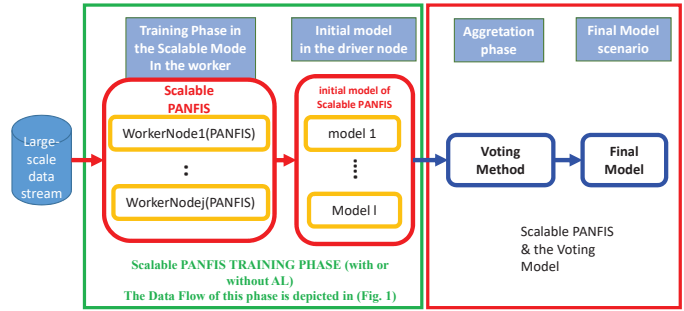}
			\par\end{centering}
		\caption{The Structure of Scalable PANFIS Framework using the Majority Voting Method\label{Structure24}}
	\end{figure}

	The initial model used in this second structure of Scalable PANFIS is the same model with the previous Scalable PANFIS framework described in subsection \ref{subsubsec:SCPANFISMerging}. However, this initial model is aggregated using the majority voting method to generate the final predictive outcomes without any merging. This structure is depicted in Fig. \ref{Structure24}.
	
	Voting methods have become popular to combine multiple classifiers,
	such as those in the early works in \cite{xu1992methods,benediktsson1992consensus,battiti1994democracy}.
	A voting method in the fuzzy rule-based classifier was pioneered
	by Ishibuchi et. al, in \cite{ishibuchi1999voting}. In the realm of fuzzy rule-based
	classifiers, there are two kinds of fuzzy rule-based voting schemes: 1) multiple fuzzy if-then
	rules in a single fuzzy rule-based classification system; and 2) multiple
	fuzzy rule-based classifiers. We adopt the second type
	majority voting where the voting procedure is carried by multiple fuzzy rule-based
	classifiers, where the voting is conducted
	in the model level instead of the rule level.
	
\begin{figure}[htbp]
		\begin{centering}
			\includegraphics[width=9cm]{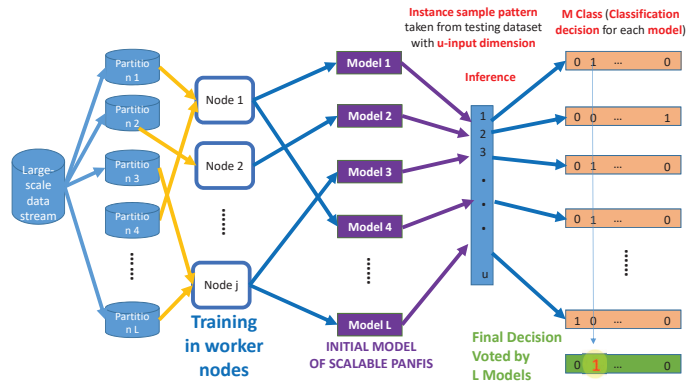}
			\par\end{centering}
		\caption{The voting mechanism scheme in the Scalable PANFIS architecture\label{votingproc}}
	\end{figure}

	PANFIS was originally designed for regression problem. For the case of classification, the output of a given $p$-instance of testing data is structured under the MIMO (Multi-Input-Multi-Output) form. The output parameter $w_i$ of $ith$-rule in the MIMO form is expressed as follows:
	
	\begin{equation}
	w_i= \left[\begin{array}{ccccc}
	w_{i0}^{1},  w_{i0}^{2}, ..., w_{i0}^{M}\\
	w_{i1}^{1},  w_{i1}^{2}, ..., w_{i1}^{M}\\
	...          ...     ...         ...	\\
	w_{iu}^{1},  w_{iu}^{2}, ..., w_{iu}^{M}\\
	\end{array}\right],
	\end{equation}
	where $M$ and $u$ denote the number of classes and input dimension, respectively. Thus, the multiplication of output parameter $w_i \in \Re^{((u+1)r) \times M}$ and
	the firing strength $\phi_{i} \in \Re^{1 \times ((u+1)r)}$ will result in the output value $y \in \Re^{1 \times M}$.
	The classification decision of a particular instance is determined by observing the highest activation degree of output
	over all rules which is expressed as follows:
	\begin{equation}
	O=\arg\max_{m=1,..,M} (O^m)%\arg\max_{m=1,...,M}(O^{m})
	\end{equation}	
	Please note that for the scalable PANFIS training process is conducted in a particular node
	of the large-scale data analytics framework which processes particular chunks/partitions of large-scale data. In the case of distributed learning, there are many partitions to be processed/learned. Therefore, the number of models generated
	depends on the initial number of partitions set before the training phase. In the case of the majority voting procedure for the final classification decision, all models generated in the large-scale data training phase influence the classification decision of every instance in the testing dataset. A visual illustration of this structure is depicted in Fig. \ref{votingproc}.
	
	The voting mechanism is applied in the Scalable PANFIS framework because it adopts 
	multi classifiers technique which often shows better performance
	than a single classifier \cite{fernandez2017fuzzy}.

	\subsubsection{Scalable PANFIS Framework with AL and the Rule Merging Method}\label{subsubsec:SCPANFISALMerging}
	This Scalable PANFIS uses distributed PANFIS with AL to  train large-scale data stream. This means that the number of samples trained in every worker node are less than number of samples in data partition processed by the worker node. From every worker node, the model is sent to the driver node as an initial model. The initial model is then merged using the rule merging method. This structure is illustrated in the Fig. \ref{Structure13}.
	
	The key feature in the large-scale data stream analytics is the
	capability of the algorithm/method to train the data efficiently in terms of running time and accuracy. In addition to speed up in the training process, the sample selection (also known as prototype reduction) is used to reduce the labeling cost because the sample evaluation procedure is done in an unsupervised mode. The sample selection in this work is inspired by the certainty-based active learning (AL) concept \cite{pratama2016incremental}. The main difference of AL from the prototype reduction method is in which AL evaluates the data in unsupervised mode.
	
	The certainty-based AL scenario is developed by virtue of the Bayesian concept, where Bayesian posterior probability determines
	the conflict level between input and output spaces following the work in \cite{pratama2016incremental}. The Bayesian
	concept is more preferable than the firing strength criterion because
	the Bayesian concept is more robust to outliers. In addition, the variable
	uncertainty strategy \cite{zliobaite2014active} counterbalances
	the effects of concept drift. This strategy adjusts the conflict threshold
	correspondingly to an up-to-date system dynamic. The substantial conflict
	in output spaces is triggered by the datum occupying an adjacent proximity
	to the decision boundary. The classifier's truncated output defines
	the conflict in the output which is expressed as follows:
	
	\begin{eqnarray}
	p(\hat{y}_{o}|X)^{output} &=& min(max(conf_{final},0),1),conf_{final} \nonumber \\
	&=&\frac{\hat{y}_{1}}{\hat{y}_{1}+\hat{y}_{2}},\label{eq:Formula10}
	\end{eqnarray}
	where $p(\hat{y}_{o}|X)^{output}$ represents the output posterior
	probability. The first and the second dominant outputs are denoted
	as $\hat{y}_{1}$and $\hat{y}_{2}$, respectively. The conflict in
	the output space is determined by the quality of the decision boundary,
	whereas the conflict of the input space is caused by the unclean cluster,
	where the cluster has a different class sample. If the classifier exhibits a strong confusion, the training samples are accepted to update the model. This criterion
	is defined as follow:
	\begin{equation}
	p(\hat{y}_{o}|X)^{output}<\theta_{randomized} \: or \: p(\hat{y}_{o}|X)^{input}<\theta_{randomized},\label{eq:Formula13}
	\end{equation}
	where $\theta$ represents the conflict threshold. Please note that $\theta$ is generated by multiplying $\theta$ with random value with the amplitude of 1 and frequency of 1 inspired by the random sampling technique \cite{zliobaite2014active}. The budget $B$
	is introduced to determine the maximum allowable number of samples to be annotated in the training process. With the assumption that data is uniformly distributed, $\theta$ is initialized as $\theta=\frac{1}{m}+B(1-\frac{1}{m})$. When the conflict in the output space or the conflict in the input space is higher than the conflict threshold, the sample does not need to be trained because a learner is confident with its own prediction. Otherwise, sample needs to be trained. The value of $\theta$ is dynamically changing, it increases when the sample is admitted for model updates, and decrease whenever sample is discarded from the training process.
	
	\subsubsection{Scalable PANFIS Framework with AL and the Majority Voting Method}\label{subsubsec:SCPANFISALVoting}
	
	The fourth structure of Scalable PANFIS framework employs distributed PANFIS with AL to train large-scale data stream to generate initial model (same initial model which is generated in subsection\ref{subsubsec:SCPANFISMerging}). This initial model contains $j$ number of classifiers (e.g. PANFIS with AL). The final model is obtained by processing the initial model using the majority voting explained in the subsection \ref{subsubsec:SCPANFISVoting}. The fourth structure of Scalable PANFIS framework (using AL) is illustrated in Fig. \ref{Structure24}.

	\section{Numerical Study }
	This section describes the numerical study of large-scale data stream analytics using Scalable PANFIS. Subsection \ref{subsec:exproc} details the experiment procedure including the environmental setup, performance measures, datasets, and methods used in the experiment. The results of the experiment are discussed in subsection \ref{subsec:result}.
	
	\subsection{Experiment Setup}\label{subsec:exproc}
	
	In this work, all the experiments are performed in the Spark platform, under the NeCTAR Cloud flexible distributed machine learning computing environment.
	The Spark platform is built by one master node and 8 worker nodes, where the NeCTAR Ubuntu 16.04 LTS (Xenial) amd64 is installed for all nodes as the operating system. Each node has the maximum specification of 390 GB disk capacity and 48GB RAM. For the total memory used in the cluster, we configure only 35GB for each worker node to be allocated in the memory cluster, leaving the rest for other operations, thus the total memory cluster is 280GB. For the driver node, we allocate 10GB for the Spark operation, leaving the rest (38GB) for other operations, considering there may be a lot of variables stored in the local memory of the driver node.
	
	In this experiment, eight algorithms are compared in order to measure their performance in terms of running time, accuracy, and the number of rules generated after the merging. The first four algorithms are the Scalable PANFIS algorithms and the other algorithms are the algorithms which are built in the Spark API library. For the sake of simplicity, we abbreviate the four structure of Scalable PANFIS algorithms which employs a combination of three techniques (e.g, the active learning, the rule merging, and the majority voting) as shown in Table \ref{algorithm}. 
	
	\begin{table}[htbp]%algorithm description
		\centering
		\caption{Algorithm description}\label{algorithm}
		\scriptsize
		\begin{tabular}{|c|c|c|}
			
			\hline
			No&\textbf{Algorithm} & \textbf{Description}   \\ \hline
			
			\multirow{ 2}{*}1&{Scalable PANFIS Merging} & Scalable PANFIS using \\
			&& \textbf{Rule Merging} Technique\\
			\hline
			
			\multirow{ 2}{*}2&{Scalable PANFIS Voting} & Scalable PANFIS using \\
			&& \textbf{Majority Voting} Technique\\
			\hline
			
			\multirow{ 2}{*}3&{Scalable PANFIS with AL Merging} & Scalable PANFIS with \textbf{AL} \\
			&& \textbf using \textbf{Rule Merging} Technique\\
			\hline
			
			\multirow{ 2}{*}4&{Scalable PANFIS with AL Voting} & Scalable PANFIS with \textbf{AL} \\
			&&  using \textbf{Majority Voting} Technique\\
			\hline
			
			5&Spark.KMeans& K-Means   \\
			\hline
			6&Spark.GLM& Spark Generalized Linear Model   \\
			\hline
			7&Spark.GBT& Spark Gradient Boosted Tree   \\
			\hline
			8&Spark.RF& Spark Random Forest   \\
			\hline
		\end{tabular}
	\end{table}
	
	For a clear explanation, Fig. \ref{Structure13} and  Fig. \ref{Structure24} show the Scalable PANFIS model sequence using the rule merging and majority voting methods as the aggregation method after initial model is generated. 
	We utilize six datasets taken from the UCI dataset repository \cite{bache2013uci} for our experiments:
	SUSY, HIGGS, HEPMASS, Poker Hand, RLCPS, and KDDCup where their specifications being shown in the Table \ref{tablesummary}. All datasets are divided into 80\% training data and 20\% testing data.
	
	Susy, Higgs, Hepmass, and RLCPS datasets are commonly used for large-scale data classification problems, such as in \cite{triguero2015mrpr,del2015mapreduce}. Poker Hand and KDDCup are multiclass datasets with 10 and 22 classes, respectively. While the KDDCup dataset features a binary classification problem: "normal" and "on attack" \cite{del2015mapreduce}.
	\begin{table}[htbp]%Dataset description
		\caption{Dataset description}\label{tablesummary}
		\begin{center}
			\begin{tabular}{|c|c|c|c|}
				\hline
				\textbf{Dataset}&\textbf{\#Sample}&\textbf{\#Atts}&\textbf{\#Class}  \\
				\hline
				SUSY&5000000  &18  &2  \\
				\hline
				HIGGS&11000000  &28  &2  \\
				\hline
				HEPMASS&10500000  &28  &2  \\
				\hline
				RLCPS&5174219  &9  &2  \\
				\hline
				Poker Hand&1025011  &10  &10  \\
				\hline
				KDDCup& 4898431 &41  &2  \\
				\hline
			\end{tabular}
		\end{center}
	\end{table}
	
	96 data partitions are set in the Scalable PANFIS framework under spark environment. Every partition is mapped into eight worker nodes to be processed in parallel mode. For each data partition, one model is generated. The number of data partitions (96 partitions) is chosen based on the consideration that our system has 96 cores with 8 driver nodes, so that every driver node processes equal number of partition (12 partitions). Increasing the number of partitions will generally speed up the training process, such as the work conducted in \cite{triguero2015mrpr}. However, our main concern in this experiment is to demonstrate the performance of Scalable PANFIS under the four big data structures.
	For each algorithm listed in Table \ref{algorithm}, four performance metrics are applied:
	\begin{enumerate}
		\item Accuracy : The percentage of correctly classified data over all testing data.
		\item Compression Rate: The performance measure of the AL method embedded in the PANFIS algorithm. This represents percentage of instances learned over all instances in the training data in particular data partitions. It also usually called compression ratio. Please note that the compression ratio is only calculated by the Scalable-PANFIS algorithm with the AL method embedded on it. Otherwise, the compression rate is 1, meaning that all samples in the dataset are used for training process.
		\item Running Time: The running time required for all
		distributed machine learning algorithms (Table \ref{algorithm}) in processing a large-scale data (Table \ref{tablesummary}) (from distributing the data partition into the worker nodes until the large-scale data model is generated in the driver node).
		\item Number of Rules: The number of rule shows the structural complexity signifying the success of the rule merging phase.
			\end{enumerate}
            Moreover, the source code of our Scalable PANFIS framework developed under R and SPARK environments are made publicly available in \footnote{\url{https://github.com/choiruzain/LargeScaleDataAnalytic}}

	\subsection{Results}\label{subsec:result}
	
	Two group of experiments are conducted to measure the performance of the algorithms. The first group compares the Scalable PANFIS without AL (Algorithm 1-2) and Scalable PANFIS with AL (Algorithm 3-4). The second group compares all the algorithms as shown in Table \ref{algorithm} (Scalable PANFIS algorithms and Spark-based algorithms). For the first group, we measure the following performance: 1)accuracy; 2)compression rate; 3)running time; 4)number of rules generated after merging. The second group measures the performance of all the algorithms (both Scalable PANFIS and Spark-based) in terms of both accuracy and running time as illustrated in Table \ref{AllAcc} and Table \ref{Allruntime}. 
	
	%The number of rules generated after merging for every dataset is illustrated in Table \ref{NumMergRule}
	
	As previously discussed, the difference between Scalable PANFIS with AL and Scalable PANFIS without AL lies in the generation of the collected model of all partitions after training phase (initial model) as shown in the Fig. \ref{Structure13} and Fig. \ref{Structure24}. Scalable PANFIS with AL only trains selected samples in each data partition, whereas Scalable PANFIS without AL trains all samples in each data partition. Therefore, it is important to compare the 
	Scalable PANFIS with AL and Scalable PANFIS without AL.
	%effect of the AL in generating the initial model, and to measure its performance in terms of the running time of the training data and its accuracy.

	\begin{table}[htbp]%Average Accuracy
		\centering
		\caption{The compression rate of the Scalable PANFIS with AL and the accuracy of Scalable PANFIS with and without AL}\label{ALcomp}
		\scriptsize
		\begin{tabular}{|c|c|c|c|}
			\hline
			\multirow{ 3}{*}{\textbf{Dataset}} & \textbf{Average} & \textbf{Average Accuracy} &\textbf{Average Accuracy}\\
			& \textbf{Compression} & \textbf{without AL(\%)} &\textbf{with AL(\%)}\\
			&\textbf{Rate(\%)}&\tiny{Merging + Voting}&\tiny{Merging + Voting}\\
			\hline
			SUSY&0.4012&76.46&\textbf{76.50}\\
			\hline
			HIGGS&0.4008&63.68&\textbf{63.82}\\
			\hline
			HEPMASS&0.4008&\textbf{83.825}&83.8\\
			\hline
			RLCPS&0.4011&\textbf{99.975}&99.935\\
			\hline
			Poker Hand&0.4008&\textbf{51.06}&48.69\\
			\hline
			KDDCup&0.408&99.73&\textbf{99.79}\\
			\hline
			
		\end{tabular}
	\end{table}
	
	Table \ref{ALcomp} shows that the average compression rate of Scalable PANFIS with AL is around 40 percent. It means that only around 40 percent of samples for each data partition are used for training purpose in the worker node on average. The average accuracy for both Scalable PANFIS with and without AL are comparable. There is no significant reduction in accuracy across all the datasets for both methods despite the reduction in the samples trained. For the SUSY, HIGGS, and KDDCup datasets, it is shown that Scalable PANFIS with AL is slightly better than Scalable PANFIS without AL. Conversely, for HEPMASS, RLCPS, and Poker Hand, Scalable PANFIS without AL slightly outperforms Scalable PANFIS with AL. From this, we can conclude that AL can operate in the training process without a loss of accuracy. Note that average accuracy is the average of classification result for Scalable PANFIS with AL or Scalable PANFIS without AL for both the rule merging and majority voting.

	A similar comparison between Scalable PANFIS with AL and without AL is shown in Table \ref{ALcomptime} in terms of running time. It is noted that the average compression rate is linear with the speed of the training partition. For example in the case of training for whole SUSY dataset, Scalable PANFIS requires 1349 seconds to generate the large-scale data model, whereas Scalable PANFIS with AL needs around half the time, this being 691 seconds. This trend is similar with the other datasets where the running time of Scalable PANFIS with AL is around half the running time of Scalable PANFIS without AL.
	
	For the number of rules generated by Scalable PANFIS with and without AL (initial model) and the number of rules after the merging is depicted in the Table \ref{NumMergRule}. It is clear that the proposed rule merging method reduces the complexity of rules in the system from before and after merging. The initial constant number of rule after merging for all datasets are constant (5 rules) as we set 5 best initial rules prior to merging process. It is proven that this method is robust to classify large-scale data stream.

	\begin{table}[htbp]%Average Running Time
		\centering
		\caption{The effect of the Active Learning Method in the Scalable PANFIS training algorithm on the running time performance}\label{ALcomptime}
		\scriptsize
		\begin{tabular}{|c|c|c|c|}
			\hline
			\multirow{ 3}{*}{\textbf{Dataset}} & \textbf{Average} & \textbf{Running Time} &\textbf{Running Time}\\
			& \textbf{Compression} & \textbf{without AL (s)} &\textbf{with AL(s)}\\
			&\textbf{Rate(\%)}&&\\
			\hline
			SUSY&0.4012&1349&691\\
			\hline
			HIGGS&0.4008&5671&2852\\
			\hline
			HEPMASS&0.4008&5280&2650\\
			\hline
			RLCPS&0.4011&630&364\\
			\hline
			Poker Hand&0.4008&218&119\\
			\hline
			KDDCup&0.4008&2965&1334\\
			\hline	
		\end{tabular}
	\end{table}

	\begin{table}[htbp]%Average Accuracy
		\centering
		\caption{Number of rule generated before and after the rule merging for initial model generated with Scalable PANFIS (with and without AL)}\label{NumMergRule}
		\scriptsize
		\begin{tabular}{|c|c|c|c|c|}
			\hline
			
			\multirow{ 3}{*}{\textbf{Dataset}} & \textbf{Scalable PANFIS} & \textbf{Scalable PANFIS} &\textbf{Scalable PANFIS}&\textbf{Scalable PANFIS}\\
			& \textbf{Before Merging} & \textbf{After Merging} &\textbf{with AL Before Merging}&\textbf{with AL After Merging}\\
			&\textbf{\# Rule}&\textbf{\# Rule}&\textbf{\# Rule}&\textbf{\# Rule}\\
			\hline
			SUSY&107&5&103&5\\
			\hline
			HIGGS&102&5&119&5\\
			\hline
			HEPMASS&135&5&137&5\\
			\hline
			RLCPS&96&5&96&5 \\
			\hline
			Poker Hand&126&5&125&5\\
			\hline
			KDDCup&96&5&96&5\\
			\hline
			
		\end{tabular}
	\end{table}

	\begin{table*}[htbp] %All accuracy for all algorithms
		\centering
		\caption{ The Accuracy for all algorithms on all datasets}\label{AllAcc}
		\begin{tabular}{|c|c|c|c|c|c|c|c|c|}
			\hline
			\multirow{2}{*}{\textbf{Algorithm}} & \multicolumn{6}{c|}{\textbf{Accuracy (\%)}}\\
			\cline{2-7}
			&\textbf{SUSY}  &\textbf{HIGGS}  &\textbf{HEPMASS}  &\textbf{RLCPS}  & \textbf{Poker Hand} &\textbf{KDDCup} \\
			\hline
			Scalable PANFIS Merging&76.70&63.66 &\textbf{83.47}&\textbf{99.98}&51.06&99.72   \\
			\hline
			Scalable PANFIS Voting&76.22& 63.70&84.18&	99.97&53.7&99.74  \\
			\hline
			Scalable PANFIS with AL Merging&\textbf{76.79}&63.72&83.45  &99.97&48.69&99.78   \\
			\hline
			Scalable PANFIS with AL Voting&76.20&\textbf{63.92}&84.15&99.9&52.8&99.82   \\
			\hline
			Spark.KMeans&50.04&	48.34&50.66&	99.63&50.21&77.21   \\
			\hline
			Spark.GLM&75.01&63.51&83.40&	99.97&50.21&78.30   \\
			\hline
			Spark.GBT&75.11&59.49&81.83&99.97&53.12& 99.88  \\
			\hline
			Spark.RF&76.81&59.65&82.43&	99.63&50.49& 99.61  \\
			\hline
		\end{tabular}
		
		\label{table2}
	\end{table*}

	Table \ref{AllAcc} and Table \ref{Allruntime} show the performance of all the algorithms in terms of accuracy, running time, respectively. In general, in terms of accuracy (Table \ref{AllAcc}), all Scalable PANFIS algorithms have a similar performance for all datasets. For example, for the SUSY, HIGGS, HEPMASS, RLCPS, Poker Hand, and KDDCup datasets, the Scalable PANFIS algorithms demonstrate an accuracy of around 76, 63, 83, 99, 51, and 99 percent, respectively. Conversely, for Spark-based 
	algorithms, the accuracy for some of the datasets are not the same. For the SUSY and HIGGS datasets, for example, Spark.KMeans algorithm is outperformed by its counterparts with only 50.04 and 48.34 percent of accuracy in comparison with other Spark-based algorithms, with around 75 and 60 percent in accuracy. Table \ref{AllAcc} also shows that for most of the datasets, Scalable PANFIS algorithms outperform Spark-based algorithms in terms of accuracy, except for the KDDCup dataset, where Spark.GBT achieves slightly better than Scalable PANFIS algorithms accuracy, while some algorithms (Spark.GLM and Spark.GBT) achieve only around 80 percent.
	
	Table \ref{Allruntime} shows that most of Spark-based algorithms perform faster when they train large-scale data. However, in one case (Poker Hand dataset), all the Spark-based algorithms (Algorithm 5-8) perform slower than the Scalable PANFIS algorithms (Algorithm 1-4).
	Of the Spark-based algorithms, Spark-GBT consumes more time than the other Spark-based algorithms. In some majority cases, Spark-GBT also performs slower than Scalable PANFIS with AL algorithms.

	\begin{table*}[htbp] %All running time for all algorithms
		\centering
		\caption{Running Time performance for all datasets and algorithms}\label{Allruntime}
		\begin{tabular}{|c|c|c|c|c|c|c|c|c|}
			\hline
			\multirow{2}{*}{\textbf{Algorithm}} & \multicolumn{6}{c|}{\textbf{Running Time (s)}}\\
			\cline{2-7}
			& SUSY & HIGGS & HEPMASS & RLCPS & Poker Hand & KDD Cup\\
			\hline
			Scalable PANFIS Merging & \multirow{2}{*}{1349} & \multirow{2}{*}{5671} & \multirow{2}{*}{5280} & \multirow{2}{*}{630} & \multirow{2}{*}{218} & \multirow{2}{*}{2965}\\
			\cline{1-1}
			Scalable PANFIS Voting &  &  &  &  &  & \\
			\hline
			Scalable PANFIS with AL Merging & \multirow{2}{*}{691} & \multirow{2}{*}{2852} & \multirow{2}{*}{2650} & \multirow{2}{*}{364} & \multirow{2}{*}{119} & \multirow{2}{*}{1334}\\
			\cline{1-1}
			Scalable PANFIS with AL Voting &  &  &  &  &  & \\
			\hline
			Spark.KMeans & 290 & 1113 & 873 & 256 & 437 & 338\\
			\hline
			Spark.GLM & 457 & 2928 & 2092 & 606 & 967 & 833\\
			\hline
			Spark.GBT & 664 & 3401 & 3167 & 852 & 1308 & 1224\\
			\hline
			Spark.RF & 259 & 1713 & 1239 & 262 & 548 & 463\\
			\hline
		\end{tabular}
	\end{table*}

	\section{Summary Discussion}\label{subsec:summdisc}
	
	This section summaries the methods, algorithms, and results obtained in this experiment. There are at least six points we can discuss in this experiment.
	
	As discussed in the results section, firstly, the AL strategy is embedded in the PANFIS machine learning algorithm to speed up the training process by selecting samples to be trained in the driver node of Scalable PANFIS framework . For many cases of large-scale data stream processing, reducing the number of samples to be trained does not have a negative effect on accuracy exemplifying the success of AL method in identifying important samples. For smaller datasets, such as Poker Hand, scalable PANFIS with AL using a merging technique yields a lower accuracy than the others (48.69 percent compared to the others). This is due to the small size of the Poker Hand dataset compared to the other datasets. With around 800k of total samples, if it is divided into 96 partitions, each partition will have around 8k samples. With the further AL applied on each partition, the samples trained in each chunk is around 3.2k (assuming the compression rate is around 0.4). Hence, the training process in all the partitions is not converged.
	
	Secondly, both the rule merging and voting techniques yield the similar performance results in terms of accuracy. The voting mechanism discards the less supported decision made by the \textbf{Weaker} classifiers which generate a false classification output, whereas the rule merging mechanism discards the classifiers which have a lower confidence level (lower weight/lower classification training results) thus resulting in better inference results. Furthermore, the over-complex rule base leads to the overfitting issue and thus deteriorates generalizing ability.
	
	Thirdly, the PANFIS architecture is designed for MIMO (multi input multi output) architecture. In the case of binary classifcation problems, the Spark-based algorithm can directly process the data. However, for the multi-class classification problem, Spark-based algorithms need to be modified into the One Versus All (OVA) form. Therefore, for the PokerHand dataset, Spark-based algorithms require a longer time to process the training data, as shown in the Table \ref{Allruntime}.
	
	Fourthly, of the Spark-based algorithms, Spark.GBT performs better than others. GBT is known as one of the most powerful techniques for building predictive models. However, it has the longest computational time of all the other Spark-based algorithms. This is because the boosting mechanism iteratively finds the suitable cost function over function space, which takes longer computational time.
	
	Fifthly, the large-scale data stream framework based on PANFIS accelerates the training process by processing all the data partitions in parallel mode. From each partition, one single model is generated
	by PANFIS with or without AL. In order to gain a final model, model fusion methods, such as rule merging and voting mechanism are applied because simply concatenating the data partition model can result in the overfitting issue deteriorating the generalization ability of the concatenated model because some rules may be overlapping.
	
	Finally, we design the robust rules merging method by selecting the initial rules which have the highest weight and applying rules removal before the rule merging process as it can be seen in the performance (accuracy and number of rules after merging). The rule merging process is explained in algorithm \ref{merging}. The rules removal is performed based on the consideration that the rules which have less support are considered as outlier, thus this could reduce the generalization capability of the classification performance. Note that our rule merging strategy is executed independently with the absence of data samples and is thus deployable for the single-pass learning context.
	
	\section{Conclusion}
	
	Evolving large-scale data stream analytics based on Scalable PANFIS demonstrates parallel data stream processing using PANFIS evolving algorithms, which combines two ways to deal with the large volume of large-scale data: streaming algorithms and distributed computing. PANFIS as an evolving algorithm can cope with data stream where its pattern is changing over time, whereas distributed computing is a platform which scales up the training phase(data process). This combination ensures that the final model generated from Scalable PANFIS (initial model) then model fusion/aggregation will ensure the large-scale data model is kept up to date. The embedded AL strategy further supports the running time in the training process of large-scale data stream analytics proven to not hit and even in some cases to improve the accuracy.
	\section{Acknowledgement}
    This work is supported by NTU Start-up Grant and MOE Tier 1 Research Grant. 
	
	\bibliography{choiru1}
	
\end{document}